\pdfoutput=1

\documentclass[11pt]{article}

\usepackage[preprint]{acl}

\usepackage{times}
\usepackage{latexsym}

\usepackage[T1]{fontenc}

\usepackage[utf8]{inputenc}

\usepackage{microtype}

\usepackage{inconsolata}

\usepackage{graphicx}
\usepackage{booktabs}
\usepackage{adjustbox}
\usepackage{hyperref}
\usepackage{tabularx}
\usepackage{longtable}
\usepackage{geometry}
\usepackage{adjustbox}
\usepackage{booktabs}
\usepackage{tablefootnote}
\usepackage{amsmath}
\usepackage{amsfonts}
\usepackage[most]{tcolorbox}
\usepackage{multicol}
\usepackage{listings}
\usepackage{enumitem}
\usepackage{authblk}
\tcbuselibrary{breakable, skins, listingsutf8}
\title{PerMedCQA: Benchmarking Large Language Models on Medical Consumer Question Answering in Persian Language}

\author{
Naghmeh Jamali\textsuperscript{1} \quad
Milad Mohammadi\textsuperscript{2} \quad
Danial Baledi\textsuperscript{2} \quad
Zahra Rezvani\textsuperscript{3} \quad
Hesham Faili\textsuperscript{2} \\
\textsuperscript{1}School of Computer Science, Institute for Research in Fundamental Sciences, Tehran, Iran. \\
\textsuperscript{2}School of Electrical and Computer Engineering, University of Tehran, Tehran, Iran. \\
\textsuperscript{3} Department of Computer Science, Faculty of Mathematical Sciences, Alzahra University, Tehran, Iran. \\
\texttt{naghme.jamali.ai@gmail.com, miladmohammadi@ut.ac.ir,} \\
\texttt{baledi.danial@gmail.com, z.rezvani@gmail.com, hfaili@ut.ac.ir}
}

\begin{document}
\maketitle
\begin{abstract}
Medical consumer question answering (CQA) is crucial for empowering patients by providing personalized and reliable health information. Despite recent advances in large language models (LLMs) for medical QA, consumer-oriented and multilingual resources—particularly in low-resource languages like Persian—remain sparse. To bridge this gap, we present \textbf{PerMedCQA}, the first Persian-language benchmark for evaluating LLMs on real-world, consumer-generated medical questions. Curated from a large medical QA forum, PerMedCQA contains 68,138 question-answer pairs, refined through careful data cleaning from an initial set of 87,780 raw entries. We evaluate several state-of-the-art multilingual and instruction-tuned LLMs, utilizing \textbf{MedJudge}, a novel rubric-based evaluation framework driven by an LLM grader, validated against expert human annotators. Our results highlight key challenges in multilingual medical QA and provide valuable insights for developing more accurate and context-aware medical assistance systems. The data is publicly available on \url{https://huggingface.co/datasets/NaghmehAI/PerMedCQA}

\end{abstract}

\section{Introduction}

Recent advances in large language models (LLMs) have significantly enhanced the capabilities of Medical Question Answering (MQA) systems, facilitating rapid and reliable access to healthcare information and supporting clinical decision-making~\cite{wang2024survey, he2025survey, zheng2025large}. These systems have demonstrated impressive performance on standardized, exam-style questions predominantly within English-language contexts, significantly aiding clinicians and patients alike by providing timely, accurate medical knowledge~\cite{meng2024application, shi2024medical, tong2025progress}. However, existing datasets and benchmarks primarily target structured, multiple-choice, or short-form question answering, often failing to capture the complexities and nuanced nature of real-world patient inquiries~\cite{pmlr-v174-pal22a, manes2024k, kim-etal-2024-medexqa}. Additionally, the emphasis on high-resource languages, particularly English, leaves substantial gaps in linguistic and cultural diversity, posing significant limitations for truly inclusive and equitable healthcare AI systems~\cite{tian-etal-2019-chimed, daoud2025medarabiq, sviridova2024casimedicos}.
\begin{figure*}[t]
  \centering
  \caption{Overview of PerMedCQA}
  \includegraphics[width=\linewidth]{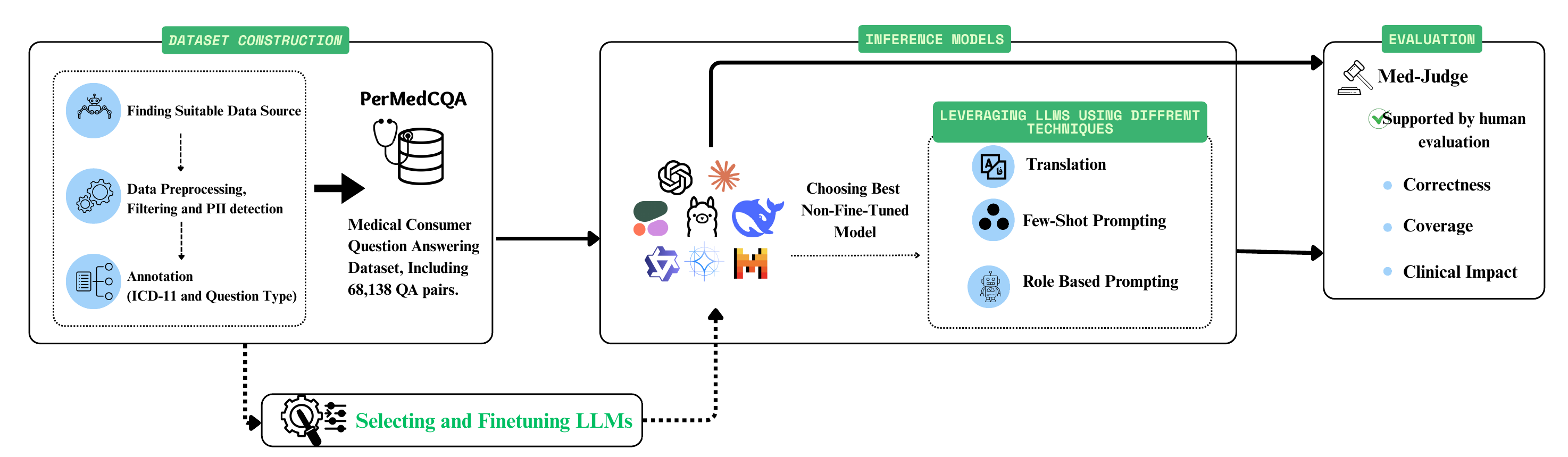}
  \label{fig:Overview of PerMedCQA}
\end{figure*}

Addressing these gaps necessitates the development of consumer-oriented Medical QA datasets that authentically reflect informal, culturally specific, and real-world patient questions encountered in everyday healthcare interactions~\cite{nguyen-etal-2023-medredqa,hosseini2024benchmark}. Such resources are particularly scarce for low-resource languages like Persian (Farsi). To the best of our knowledge, no existing studies or datasets explicitly focus on Persian medical consumer question answering, highlighting a critical barrier to the advancement of equitable and culturally-aligned healthcare AI systems in this language.

To bridge this critical gap, we introduce \textbf{PerMedCQA}, the first large-scale Persian-language benchmark specifically designed for consumer-oriented medical question answering. PerMedCQA comprises 68,138 carefully curated question-answer pairs derived from real-world interactions collected from four prominent Persian online medical forums. These interactions involve authentic consumer-generated health queries answered by licensed medical professionals, encompassing a wide range of medical specialties and enriched with detailed metadata such as patient demographics, physician specialty, and platform source.

To ensure the dataset's reliability, clinical utility, and compliance with privacy standards, we implemented a rigorous two-stage data cleaning process. Initially, rule-based heuristics were applied to remove invalid entries, extremely short interactions, duplicates, and non-textual content, significantly enhancing the data's structural quality. Subsequently, we employed a LLM to label instances containing personally identifiable information (PII) such as names, phone numbers, addresses, and emails, further ensuring user privacy and ethical compliance. This comprehensive preprocessing pipeline resulted in a high-quality, de-identified dataset suitable for robust medical AI research.

To facilitate rigorous benchmarking and fine-grained analyses, we systematically annotated each QA pair along two critical dimensions: (1) disease categorization using the International Classification of Diseases, 11th Revision (ICD-11)~\cite{Khoury2017}, comprising 28 distinct medical categories; and (2) classification into one of 25 standardized question types that capture the structural and semantic intent of patient inquiries~\cite{abacha2019bridging}. This extensive annotation significantly enhances PerMedCQA's utility for comprehensive evaluation and analysis of medical LLM capabilities.

Given the inherent challenges in evaluating open-ended medical QA tasks, where clinically acceptable responses often vary significantly in phrasing, we adopt an innovative evaluation framework utilizing a large language model (\textbf{Med-Judge}) as an automated grader. Validated through expert physician assessments, this rubric-driven system systematically compares model-generated responses against expert-provided answers, enabling nuanced evaluations beyond traditional lexical similarity metrics like BLEU or ROUGE.

We conducted extensive benchmarking experiments across diverse state-of-the-art language models, including both proprietary and prominent open-source variants. Robust baseline performance was established through zero-shot inference, followed by exploration of several inference-time enhancements—such as role-based conditioning and pivot translation—to improve response quality without parameter updates. Additionally, supervised fine-tuning experiments using parameter-efficient methods (LoRA) were performed on selected smaller models to assess the learnability and practical utility of PerMedCQA as a training resource.

Our key contributions can be summarized as follows:
\begin{itemize}
    \item We introduce \textbf{PerMedCQA}, the first large-scale, real-world Persian medical QA benchmark, meticulously constructed through a rigorous two-step data cleaning process—rule-based filtering and LLM-based PII detection—and annotated comprehensively with ICD-11 categories and standardized question types, substantially addressing the resource gap for low-resource language medical QA while ensuring data quality, privacy, and ethical compliance.
    \item We present an automated yet clinically informed evaluation framework (\textbf{Med-Judge}), validated by expert physician reviews, providing reliable and nuanced assessments of open-ended medical question answering quality.
    \item We comprehensively benchmark a variety of proprietary and open-source LLMs, identifying substantial performance variations and demonstrating the effectiveness of prompt-based techniques for enhancing model outputs.
    \item We evaluate supervised fine-tuning strategies using PerMedCQA, highlighting their potential and limitations for improving smaller-scale model performance and clinical reliability.
\end{itemize}

An overview of the complete PerMedCQA workflow—from data collection and annotation through to model evaluation—is provided in Figure~\ref{fig:Overview of PerMedCQA}. Through these contributions, PerMedCQA establishes a critical foundation for future research aimed at developing trustworthy, culturally-sensitive, and linguistically inclusive medical AI systems specifically tailored to Persian-speaking populations.

\section{Related Work}
The rapid development of Large Language Models (LLMs) has significantly advanced the field of Medical Question Answering (Medical QA). While models such as GPT-4~\cite{nori2023capabilities}, PaLM~\cite{chowdhery2023palm}, Mistral~\cite{chaplot2023albert}, and LLaMA~\cite{touvron2023llama} have achieved impressive results on English-language Medical QA benchmarks, the progress of LLMs in non-English and consumer-focused medical domains remains underexplored. In this section, we review key datasets and modeling approaches that have shaped current research in Medical QA.

\subsection{Medical Question Answering Dataset}
The progress in Medical QA is closely tied to the availability of high-quality and diverse datasets. To evaluate clinical accuracy and factual consistency, LLMs have been tested on exam-style multiple-choice benchmarks such as MedQA~\cite{jin2021disease} and PubMedQA~\cite{jin2019pubmedqa}. These benchmarks feature questions framed in the style of medical licensing exams, focusing on factual recall. However, multiple-choice formats often fail to reflect the complexity and nuance of real-world medical inquiries, as they constrain responses to predefined options and limit models’ ability to generate explanatory or contextual answers~\cite{welivita2023survey}. To address these limitations, recent benchmarks such as Medical Long-Form QA~\cite{hosseini2024benchmark} and MedRedQA~\cite{nguyen-etal-2023-medredqa} prioritize practical utility and open-ended responses, aligning more closely with consumer-oriented healthcare needs.

While most Medical QA datasets are in English, efforts have been made to extend coverage to other languages. For Chinese, ChiMed~\cite{tian-etal-2019-chimed}, built from large online medical forums, serves as a benchmark for QA in Chinese. MedDialog~\cite{zeng-etal-2020-meddialog}, comprising real-world, open-ended medical dialogues in both Chinese and English, has enabled progress in conversational medical systems via transfer learning. Huatuo-26M~\cite{li2023huatuo}, a large-scale dataset with 26 million QA pairs sourced from Chinese encyclopedias, knowledge bases, and online consultations, has further boosted model training. More recently, MMedC~\cite{qiu2024towards} was introduced as a multilingual medical corpus containing approximately 25.5 billion tokens across six languages, aiming to support the development of more capable and generalizable medical LLMs, particularly for low-resource languages. For Arabic, AraMed~\cite{alasmari-etal-2024-aramed} provides QA pairs extracted from AlTibbi, an online doctor-patient discussion platform.

Several foundational datasets have been consolidated into comprehensive benchmarks for evaluating LLMs in medical QA.~\cite{jin2019pubmedqa, ben-abacha-etal-2019-overview, jin2021disease} MultiMedQA~\cite{singhal2023publisher}, for instance, aggregates six multiple-choice datasets—MedQA (USMLE), MedMCQA~\cite{pal2022medmcqa}, PubMedQA, LiveQA~\cite{abacha2019overview}, MedicationQA~\cite{abacha2019bridging}, and MMLU clinical topics~\cite{hendrycks2020measuring}—along with the HealthSearchQA~\cite{singhal2023large} dataset, to support a wide range of evaluation tasks. In the context of Retrieval-Augmented Generation (RAG), MIRAGE~\cite{xiong2024benchmarking} offers a standardized benchmark that combines subsets of MedQA, MedMCQA, PubMedQA, MMLU clinical topics, and BioASQ-YN~\cite{tsatsaronis2015overview}, enabling systematic evaluation of RAG techniques in medical QA.

\subsection{Medical Large Language Models (LLMs in healthcare)}
Recent years have seen the emergence of several domain-specific LLMs for healthcare, such as Med-Gemini~\cite{saab2024capabilities}, Med-PaLM~\cite{singhal2023publisher}, MedPaLM-2~\cite{singhal2025toward}, BioMistral~\cite{labrak-etal-2024-biomistral}, PMC-LLaMa~\cite{wu2024pmc} and MMed-Llama 3~\cite{qiu2024towards}, each demonstrating superior performance in medical reasoning and generation tasks. These advancements necessitate the development of more robust and nuanced benchmarks to thoroughly assess model capabilities.

The field has also expanded with fine-tuned medical LLMs such as HuatuoGPT~\cite{zhang2023huatuogpt}, BianQue~\cite{chen2023bianque}, ClinicalGPT~\cite{wang2023clinicalgpt}, DoctorGLM~\cite{xiong2023doctorglm}, Chatdoctor~\cite{li2023chatdoctor}, Baize-healthcare~\cite{xu2023baize}, zhongjing~\cite{yang2024zhongjing}, Clinical Camel~\cite{toma2023clinical}, and Me-LLaMA~\cite{xie2024me}, many of which were trained on real-world clinical notes and patient-doctor dialogues. In parallel, multilingual medical models such as CareBot~\cite{zhao2025carebot}, Medical-mT5~\cite{garcia2024medical}, ChiMed-GPT~\cite{tian-etal-2024-chimed}, and BiMediX~\cite{pieri-etal-2024-bimedix} have extended support for cross-lingual medical applications, increasing the global accessibility of medical AI systems.

Despite these advancements, There is a gap in the literature when it comes to real-world datasets and most open-ended Medical QA datasets remain limited to English, leaving a substantial gap in multilingual evaluation—especially for low-resource languages. 
\section{PerMedCQA Dataset Construction}
To support the development and evaluation of medical consumer question answering systems in Persian, we introduce PerMedCQA — a large-scale dataset of real-world consumer health questions and expert answers, sourced from verified specialists across multiple public Persian forums. This section details the data collection process, preprocessing pipeline, automatic annotation, and benchmark construction. 

\subsection{Data Sources and Raw Collection}

The initial dataset comprises 87,780 question-answer pairs collected from four major Persian-language health Q\&A platforms: \cite{DrYab}, \cite{HiSalamat}, \cite{GetZoop}, and \cite{Mavara-e-Teb}. Each platform hosts a diverse set of verified physicians, collectively covering over 100 medical specialties. All data was collected from publicly accessible pages between November 10, 2022, and April 2, 2024, in compliance with ethical standards for public web data usage.

Each QA instance includes metadata fields such as \texttt{Title}, \texttt{Category} (user-assigned), \texttt{Physician Specialty}, \texttt{Age}, and \texttt{Sex}. All metadata across sources were standardized to a unified format, ensuring consistency. The dataset includes various QA interaction structures: (1) single-turn dialogues (DrYab, HiSalamat), (2) multi-turn conversations (GetZoop), and (3) cases with multiple expert responses (Mavara-e-Teb).

\subsection{Preprocessing and Cleaning}

We employed a two-stage data cleaning pipeline: rule-based preprocessing and large language model (LLM)--based processing (filtering and tagging).

\paragraph{Rule-based Filtering.} 
Briefly, the following heuristics were applied across all data sources: (1) Entries without a valid user or assistant message were removed. (2) QA pairs where either message had fewer than three words, images, videos, or URLs were discarded. (3) Duplicate (user, assistant) pairs across different files were eliminated. These rule-based filters reduced the dataset from 87,780 to 73,416 instances, removing a total of 14,364 entries.

\paragraph{PII Detection.} 
Given the sensitive nature of healthcare data, we further removed QA pairs containing personally identifiable information \cite{PII}, such as names, phone numbers, addresses, and emails. GPT-4o-mini was employed to detect any records containing PII with high accuracy. This stage resulted in the removal of 5,278 additional instances. The final cleaned dataset contains 68,138 QA pairs, resulting the final numbers QA pairs in PerMedCQA. Table \ref{tab:Disrtibution of Qa pairs in the data sources} shows the distribution of QA pairs in terms of their data source. 
\begin{table}[h!]
\centering
\caption{Distribution of QA pairs in the data sources}
\begin{tabular}{ccc}
\hline
\textbf{Data Source} & \textbf{\#QA Pairs} & \textbf{Percentage} \\
\hline
DrYab & 34427 & 50.5\% \\
GetZoop & 24006 & 35.2\% \\
HiSalamat & 5011 & 7.4\% \\
Mavara-E-Teb & 4692 & 6.9\% \\
\hline
\end{tabular}
\label{tab:Disrtibution of Qa pairs in the data sources}
\end{table}
\subsection{Annotation and Benchmark Splitting}

\paragraph{ICD-11 Tagging.} \cite{Khoury2017}
To facilitate medically meaningful categorization, the International Classification of Diseases 11th Revision (ICD-11) that consists of 28 categories, assigned to each QA pair by employing GPT-4o-mini. The prompt for ICD-11 classification and PII tagging showed in Figure \ref{fig:Instructions for ICD-11 classification and PII tagging} in Appendix \ref{sec:appendix}. This process yielded 28 distinct ICD-11 classes. Unlike user-assigned categories, ICD-11 provides a consistent and standardized taxonomy for disease classification. Figure \ref{fig:Distribution-ICD-11} shows the distribution of ICD-11 categories in PerMedCQA.

\paragraph{Question Type Tagging.}
We categorized each QA pair into one of 25 predefined question types~\cite{abacha2019bridging}, enabling a deeper structural understanding of the dataset. The definition of each category illustrates in Table \ref{fig:Question Type Categories} along with some examples and the prompt of question type tagging is shown in Figure \ref{fig:Question Type Tagging Task-25}. Figure \ref{fig:QuestionType Distribution} shows the distribution of ICD-11 categories in PerMedCQA.
\begin{figure}[t]
    \centering
    \caption{Gender Distribution in PerMedCQA}
    \includegraphics[width=\linewidth]{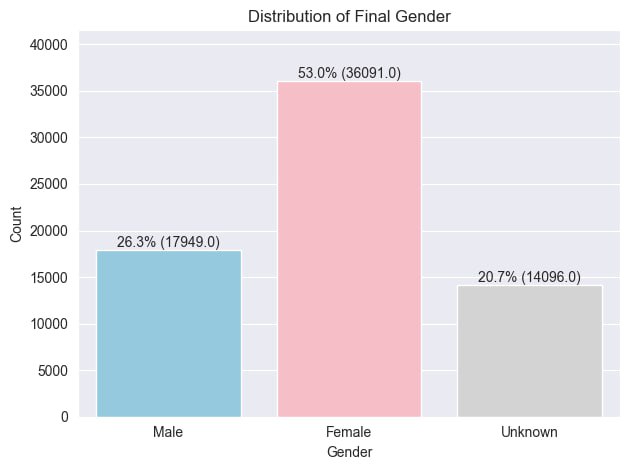}
    \label{fig:Gender Distribution in PerMedCQA}
\end{figure}

\paragraph{Extra Analysis based on Gender.}
Culture affects shaping the distribution of questions across ICD-11 disease categories \ref{fig:Distribution-ICD-11}, question types \ref{fig:QuestionType Distribution}, and the gender of questioners. As shown in Figure \ref{fig:Gender Distribution in PerMedCQA}, the gender-based distribution indicates that women and unknown constitute the majority of users who post questions on Persian Forums. The highest number of questions are allocated to the categories of sexual health (9,266 questions), digestive systems (6,868 questions), and Skin care and disease (5,478 questions), comprising roughly one-third of the entire dataset. In all three categories, the dominant "gender" of the questioners is "women", and the "question type" is predominantly "information". This observation highlights the substantial cultural influence on the prominence of these categories. For example, the tendency among women to ask about their sexual health concerns anonymously, seek home remedies for digestive issues, and obtain skincare advice without in-person consultation reflects cultural norms, explaining the higher participation of women in Persian-language medical forums.

\paragraph{Benchmark Split.}
To support model training and evaluation, PerMedCQA were split according to appropriate percent subsets for training, evaluation and test, partitioning into: Train set contains 64,280 instances, Eval set includes 345 instances (15 per ICD-11 category), and Test set comprises 3,513 instances (150 per ICD-11 category). Evaluation and test splits were stratified by ICD-11 category to ensure balanced representation across medical domains.

\section{Experiments}
This section outlines the experimental setup used to evaluate models on the PerMedCQA dataset. Due to the long-form (LF)  nature of the task, we adopt a structured evaluation approach using a large language model as an automatic judge \cite{zheng2023judgingllmasajudgemtbenchchatbot}, referred to as Med-Judge, used to assess model performance on the PerMedCQA dataset. To validate the reliability of this automatic evaluation protocol, we also conducted a human expert evaluation like \cite{hosseini2024benchmark}. We then benchmarked a range of state-of-the-art language models using zero-shot inference to establish robust baselines. Based on this initial evaluation, the best-performing model in Persian for further analysis was chosen to apply a variety of advanced inference techniques (e.g, prompt-based strategy). Furthermore, to assess the quality and learnability of the PerMedCQA dataset, supervised fine-tuning (SFT) was applied to a set of language models. 
\subsection{LLM–based Evaluation with \texttt{Med-Judge}}
Traditional evaluation metrics such as BLEU~\cite{papineni-etal-2002-bleu}, ROUGE~\cite{lin-2004-rouge}, METEOR~\cite{banerjee-lavie-2005-meteor} penalize legitimate paraphrases and thus fail for open-ended medical QA, where multiple correct answers may differ significantly in wording while still conveying clinically equivalent information. To address this limitation, we adopted \textbf{\texttt{Med-Judge}}, a LLM \cite{Gemini-Flash-2.5}~\cite{team2024gemini}) prompted to evaluate model answers against expert references using a predefined rubric. Med-Judge is based on the criteria \cite{hosseini2024benchmark}, comprising Correctness (Correct, Partially correct, Incorrect, Contradictory), Coverage (equal, model\_subset, expert\_subset, overlap\_none), and Critical\_Impact (Negligible, Moderate, Significant, Critical), explaining further details in \ref{sec:appendix-medjudge}. Our full prompt of Med-Judge is shown in Figure \ref{fig:MedJudge system prompt} and \ref{fig:User prompt} in section \ref{sec:appendix}. 
\subsection{Baseline Model Benchmarking}
To establish robust performance benchmarks on the PerMedCQA dataset, we conducted zero-shot evaluations as an initial evaluation across a diverse suite of both proprietary and open-source language models. In total, we tested 16 models spanning a range of scales and training paradigms, primarily multilingual general-purpose models, along with a single biomedical-tuned variant. All evaluations were conducted on the fixed PerMedCQA test set, allowing for consistent comparisons across methods. The full list of evaluated models is summarized in Table \ref{tab:List of LLM models in our experiments}.
\begin{table}[h!]
\centering
\small
\caption{The list of LLM models in our experiments}
\begin{tabular}{c|c}
\hline
\textbf{Model Name} &  \textbf{Affiliation} \\
\hline
GPT-4.1, 4.1-mini, 4.1-nano \\ ~\cite{achiam2023gpt}& OpenAI \\
\hline
Claude 3.5 Haiku~\cite{Claude3.5Haiku} & Anthropic \\
\hline
Claude 3.7 Sonnet~\cite{Claude3.7Sonnet} & Anthropic \\
\hline
Mistral-Saba~\cite{mistral-saba} & Mistral \\
\hline
Command A- 111B~\cite{cohere2025command} & Cohere \\
\hline
BioMistral-7B~\cite{labrak2024biomistral}& Mistral \\
\hline
DeepSeek-V3-670B~\cite{liu2024deepseek} & DeepSeek \\
\hline
LLaMA 4 Scout-109B~\cite{LLaMA4Scout-109B} & Meta \\
\hline
LLaMA 3.3 70B & Meta \\
\hline
LLaMA 3.1 8B~\cite{grattafiori2024llama}& Meta \\
\hline
Gemma3~\cite{gemma_2025} 4B, 12B, 27B & DeepMind \\
\hline
qwen3-14b~\cite{yang2025qwen3technicalreport} & Qwen \\
\hline
\end{tabular}
\label{tab:List of LLM models in our experiments}
\end{table}

All models baseline experiments were evaluated under a consistent zero-shot setting using the same prompt. This prompt instructed the model to adopt the persona of a professional medical doctor answering in fluent Persian. The instructions emphasized direct, precise, and actionable guidance, discouraging excessive elaboration or default referrals unless medically necessary. Prompt structure and a representative example of the prompt is shown in Figure \ref{fig:Baseline system prompt} in Appendix \ref{sec:appendix}. 

\subsection{Prompt-based Enhancement Methods}
In addition to the baseline zero-shot evaluations, inference-time and non-parametric strategies were explored to aim the improvement of model performance without gradient updates or fine-tuning. These techniques were designed to enhance answer quality by modifying the input context or interaction style, while keeping the model parameters fixed. The three main techniques considered are described in the following paragraphs:

\paragraph{Pivot Translation for LLM Processing.}~\cite{tanaka2024performance} 
To leverage the strength of LLMs in English, we translated the Persian input question into English, requested an English-language response from the model, and then performed back-translation the answer to Persian. GPT-4.1 was employed for both directions, the prompt shows in Figure \ref{fig:Translation pipeline} in \ref{sec:appendix}. This method introduces latency but often enhances fluency, completeness, and structured reasoning, potentially compensating for model weaknesses in low-resource language handling. 

\paragraph{Role-based Prompting.}\cite{grabb2023impact}
Each question was prepended with a system-level role prompt tailored to its ICD-11 category. For example, if the ICD-11 tag was related to mental disorders (tag 6), the model received the instruction: \textit{"You are an experienced 'Psychiatrist' providing reliable..."}. This approach aimed to inject domain-specific priors and guide the model toward more specialized, context-aware answers. The prompt shows in Figure \ref{fig:Role-based prompting strategy} in \ref{sec:appendix}. 

\paragraph{Few-shot Prompting.}\cite{maharjan2024openmedlm}
To test few-shot prompting, where each model was shown five randomly selected QA pairs from the PerMedCQA training set before answering the target question. The exemplars were selected without regard to topic simila
rity to preserve generalizability. This strategy aimed to improve format consistency, response completeness, and adherence to clinical style by offering implicit demonstration of expected output structure. These techniques were applied to the best-performing model from the baseline evaluation, selected based on Med-Judge scores. Their effects discussed in Section \ref{sec:Results and Analysis}.

\subsection{Supervised Fine-Tuning}
To assess the learnability and utility of the PerMedCQA dataset, supervised fine-tuning (SFT) experiments were conducted on three language models, comprising Gemma 4B, LLaMA 3.1 8B, and BioMistral 7B. In terms of accessibility, architectural diversity, and suitability for efficient instruction tuning in resource-constrained environments, these models were selected. The training was conducted on the full PerMedCQA training set for 1 epoch, comprising 64,279 QA pairs. Moreover, despite lacking native support for Persian, BioMistral was included due to its specialization in biomedical and clinical domains. We employed LoRA-based (Low-Rank Adaptation) parameter-efficient fine-tuning~\cite{hu2022lora}, using LLaMaFactory framework \cite{zheng2024llamafactory}. LoRA configuration included a rank of 8 and $\alpha = 16$, with a learning rate of $2 \times 10^{-5}$ and context length of 2048 tokens. 

Rather than aiming for state-of-the-art performance, these fine-tuning experiments were primarily designed to evaluate the effectiveness of PerMedCQA as a training resource, and to explore how much medical QA capability could be learned by modestly sized models through supervised instruction tuning.

\section{Results and Analysis and Conclusion}
\label{sec:Results and Analysis}
This section is currently under development and will be included in a future revision.
\section*{Limitations}
Despite the comprehensive scale and rigorous cleaning of PerMedCQA, our study faces several limitations. The dataset, while large, is derived from a limited set of public Persian medical forums, which may introduce topical and demographic biases and restrict generalizability beyond Persian-speaking communities or to clinical domains not well represented in our sources. Furthermore, while we implemented robust quality control—including automated PII removal and human expert validation—our human evaluation was limited by the availability of clinical experts, restricting the depth and diversity of manual review. Finally, the dataset focuses on text-only QA, excluding cases requiring visual information, which are common in real-world healthcare settings and present important avenues for future work.

\section*{Ethical Considerations}
Throughout this work, we prioritized privacy, fairness, and legal compliance. All data were collected exclusively from publicly accessible forums, and a thorough multi-stage pipeline was applied to remove any personally identifiable information (PII), ensuring no private patient or clinician data is present in the released dataset. We confirmed that data collection practices respected the terms and conditions of the source websites, and we will respond to takedown requests as needed. While releasing PerMedCQA, we recognize the risk of perpetuating cultural or topical biases inherent in consumer-generated content, as well as the potential for LLMs to generate incorrect or misleading medical advice. As such, both the dataset and any models fine-tuned with it are intended strictly for research purposes and should not be used as a substitute for professional medical care. We strongly recommend that downstream applications include clear disclaimers, robust safety measures, and, when possible, human expert oversight to prevent harm from inaccurate or inappropriate responses.

\bibliography{PerMedCQA}

\appendix
\section{Appendix}
\label{sec:appendix}

\subsection{Med\_Jugde}
\label{sec:appendix-medjudge}
For every test instance, the MedJudge LLM receives the user’s question, the expert-provided “gold” answer, and the model-generated answer. The LLM is strictly instructed to evaluate based solely on the expert reference, without drawing from external sources or its own medical knowledge. The output is a structured JSON object with the following fields:

\begin{itemize}
    \item \textbf{Brief\_analysis}: A concise comparison highlighting key similarities or differences between expert and model answers.
    \item \textbf{Key\_missing\_facts}: Important medical facts present in the expert answer but missing from the model answer (in Persian).
    \item \textbf{Key\_extra\_facts}: Additional details present in the model answer but absent from the expert answer (in Persian).
    \item \textbf{Correctness}: Assessment of factual and clinical consistency: \textit{Correct}, \textit{Partially\_correct}, \textit{Incorrect}, or \textit{Contradictory}.
    \item \textbf{Coverage}: Degree of factual overlap: \textit{Equal}, \textit{Model\_subset}, \textit{Expert\_subset}, or \textit{Overlap\_none}.
    \item \textbf{Clinical\_impact}: Estimated clinical significance of any discrepancies: \textit{Negligible}, \textit{Moderate}, \textit{Significant}, or \textit{Critical}.
    \item \textbf{Judge\_confidence}: The LLM’s self-reported confidence (\textit{High}, \textit{Medium}, \textit{Low}).
\end{itemize}

\subsection{Label Definitions, Scoring, and Reliability}
\paragraph{Label Definitions}
Each output field is defined as follows:

\begin{itemize}
    \item \textbf{Correctness}: 
        \begin{enumerate}
            \item \textit{Correct}: Clinically equivalent; no meaningful differences.
            \item \textit{Partially\_correct}: Minor deviations, no significant clinical impact.
            \item \textit{Incorrect}: Substantial differences that affect accuracy or completeness.
            \item \textit{Contradictory}: Model advice directly conflicts with the expert reference.
        \end{enumerate}
    \item \textbf{Coverage}:
        \begin{enumerate}
            \item \textit{Equal}: Both answers contain the same key facts.
            \item \textit{Model\_subset}: Model omits critical facts present in the expert answer.
            \item \textit{Expert\_subset}: Model introduces relevant facts not found in the expert answer.
            \item \textit{Overlap\_none}: No substantial factual overlap.
        \end{enumerate}
    \item \textbf{Clinical\_impact}:
        \begin{enumerate}
            \item \textit{Negligible}: No effect on care or understanding.
            \item \textit{Moderate}: Slight effect on treatment or comprehension.
            \item \textit{Significant}: Likely to affect recommendations or outcomes.
            \item \textit{Critical}: May result in unsafe or harmful guidance.
        \end{enumerate}
\end{itemize}

\paragraph{Reliability}

To assess the reliability of MedJudge, we compared its labels against blinded ratings from board-certified physicians on a 100-item subset. Agreement on the primary dimension—\textit{correctness}—was 75\% (collapsed to ``acceptable'' vs. ``problematic''), with quadratic Cohen’s~$\kappa = 0.42$ (95\% CI 0.19–0.58) and $F_{1} = 0.82$. Ordinal rank-correlation was significant ($\tau = 0.41$, $p < 0.001$), indicating that higher human scores were mirrored by the LLM. For \textit{clinical-impact}, MedJudge achieved 78\% accuracy and detected one-third of truly high-impact discrepancies ($\kappa = 0.07$; prevalence-adjusted).
\subsection{Prompt Engineering}
\begin{figure*}[t]
  \centering
  \caption{MedJude Structured Output Schema}
  \includegraphics[width=\linewidth]{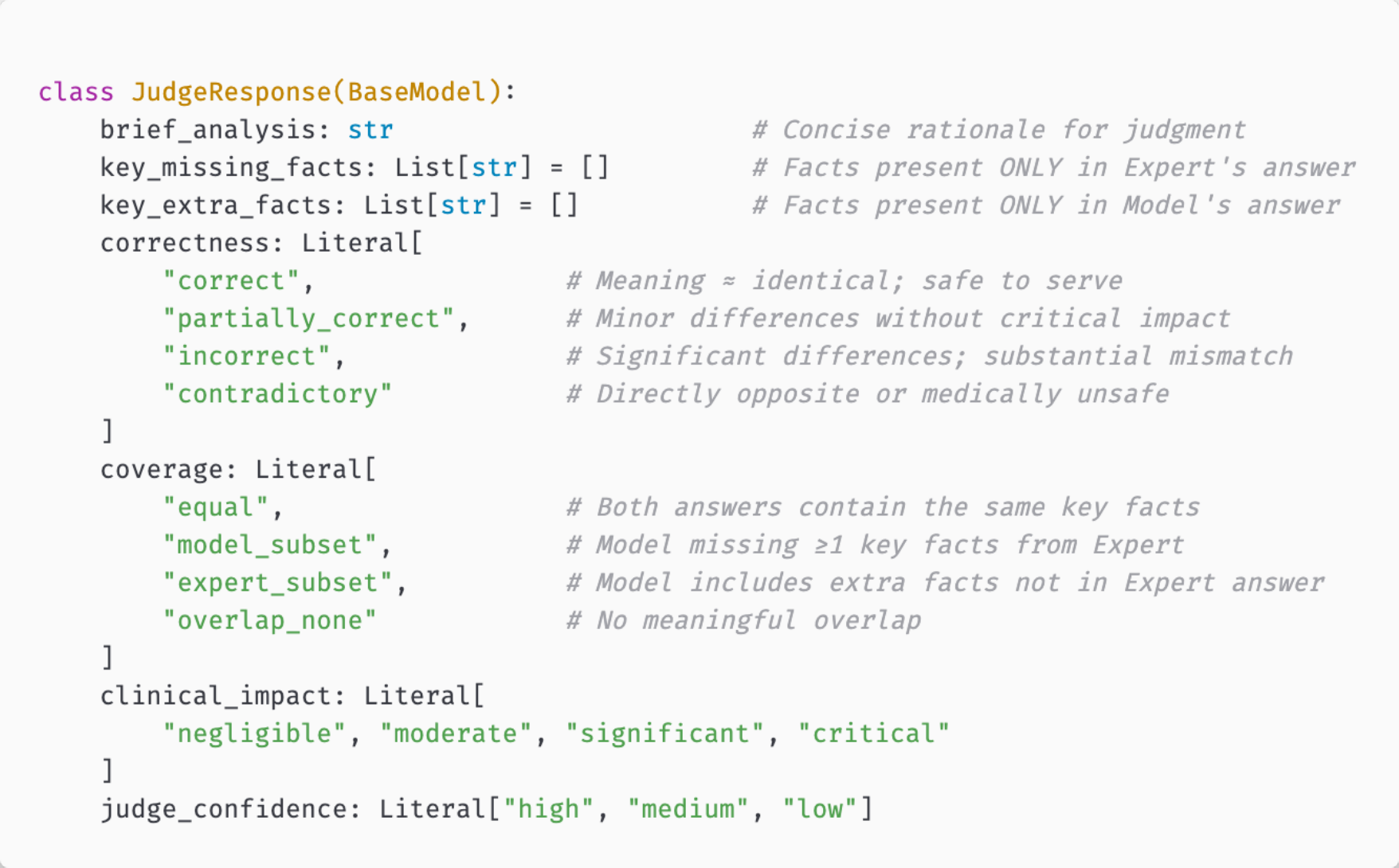}
  \label{fig:MedJudgeSchema}
\end{figure*}

Prompt engineering is considered crucial for the evaluation of LLMs. In this section, the prompts used at various stages of the work are presented. Figures~\ref{fig:Instructions for ICD-11 classification and PII tagging} and~\ref{fig:icd11-catalogue} illustrate the task instructions provided for ICD-11 classification and PII tagging during the MedPerCQA stage. As shown in Figure~\ref{fig:Question Type Tagging Task-25}, question type tagging across 25 categories was performed using a dedicated prompt. In addition, a wide range of prompts was employed in extensive experiments, as illustrated in Figures~\ref{fig:Baseline system prompt},~\ref{fig:Role-based prompting strategy}, and~\ref{fig:Translation pipeline}. Finally, the structured output schema used in the Med-Judge evaluation pipeline is shown in Figure~\ref{fig:MedJudgeSchema}, enabling reliable parsing and analysis of model responses; the prompt design incorporates a chain-of-thought strategy to guide the LLM through consistent reasoning steps.

\begin{figure*}[t]
\caption{Task instructions for ICD-11 classification and PII tagging.}
\begin{tcolorbox}[
    enhanced,
    width=\textwidth,
    colback=gray!5,
    colframe=blue!75!black,
    title=\textbf{ICD-11 and PII Tagging System Prompt},
    fonttitle=\bfseries,
    sharp corners=south,
    boxrule=0.6pt
  ]
\small

\textbf{Task Description} \\
You are a medical expert tasked with classifying and analyzing patient–expert dialogue.  
Your task has two main parts:

\bigskip
\textbf{1) ICD-11 Classification} \\
Classify the content based on ICD-11 categories using the catalogue provided in the next page.  
Return only the integer (1–28) that best corresponds to the core subject matter.

\bigskip
\textbf{2) PII Detection} \\
Check if any personal information of the patient or expert is exposed in the messages.  
Personal information includes examples such as real name, address, phone number, email, etc.  
If such information exists, set \texttt{"identity"} to \texttt{true}; otherwise, set it to \texttt{false}.

\end{tcolorbox}
\label{fig:Instructions for ICD-11 classification and PII tagging}
\end{figure*}
\begin{figure*}[t]
  \centering
  \caption{Standardized ICD-11 classification codes used for QA annotation.}
  \begin{tcolorbox}[
    enhanced,
    width=\textwidth,
    colback=gray!5,
    colframe=gray!75!black,
    title   = \textbf{ICD-11 Category Catalogue (28-class taxonomy)},
    fonttitle=\bfseries,
    sharp corners=south,
    boxrule = 0.6pt
  ]
  \small
  \begin{multicols}{2}
  \begin{enumerate}[leftmargin=*]
    \item (1A00–1H0Z) Certain infectious or parasitic diseases
    \item (2A00–2F9Z) Neoplasms
    \item (3A00–3C0Z) Diseases of the blood or blood-forming organs
    \item (4A00–4B4Z) Diseases of the immune system
    \item (5A00–5D46) Endocrine, nutritional or metabolic diseases
    \item (6A00–6E8Z) Mental, behavioural or neurodevelopmental disorders
    \item (7A00–7B2Z) Sleep-wake disorders
    \item (8A00–8E7Z) Diseases of the nervous system
    \item (9A00–9E1Z) Diseases of the visual system
    \item (AA00–AC0Z) Diseases of the ear or mastoid process
    \item (BA00–BE2Z) Diseases of the circulatory system
    \item (CA00–CB7Z) Diseases of the respiratory system
    \item (DA00–DE2Z) Diseases of the digestive system
    \item (EA00–EM0Z) Diseases of the skin
    \item (FA00–FC0Z) Diseases of the musculoskeletal system or connective tissue
    \item (GA00–GC8Z) Diseases of the genitourinary system
    \item (HA00–HA8Z) Conditions related to sexual health
    \item (JA00–JB6Z) Pregnancy, childbirth or the puerperium
    \item (KA00–KD5Z) Certain conditions originating in the perinatal period
    \item (LA00–LD9Z) Developmental anomalies
    \item (MA00–MH2Y) Symptoms, signs or clinical findings, not elsewhere classified
    \item (NA00–NF2Z) Injury, poisoning or other consequences of external causes
    \item (PA00–PL2Z) External causes of morbidity or mortality
    \item (QA00–QF4Z) Factors influencing health status or contact with health services
    \item (RA00–RA26) Codes for special purposes
    \item (SA00–SJ3Z) Traditional Medicine Conditions – Module I
    \item (VA00–VC50) Functioning assessment
    \item (XA0060–XY9U) Extension Codes
  \end{enumerate}
  \end{multicols}
  \end{tcolorbox}
  \label{fig:icd11-catalogue}
\end{figure*}
\begin{table*}[t]
\centering
\caption{Question Type Categories.}
\begin{tabular}{p{0.18\linewidth}|p{0.42\linewidth}|p{0.32\linewidth}}
\hline
\textbf{Question Type} & \textbf{Definition} & \textbf{Example} \\
\toprule
Information & Asks for general identification or classification of a drug. & What type of drug is amphetamine? \\
\hline
Dose & Queries recommended or safe dosage. & What is a daily amount of prednisolone eye drops to take? \\
\hline
Usage & Seeks instructions on how to take/administer a drug. & How to self inject enoxaparin sodium? \\
\hline
Side Effects & Asks about adverse reactions. & Does benazepril aggravate hepatitis? \\
\hline
Indication & Asks why/for which condition the drug is prescribed. & Why is pyridostigmine prescribed? \\
\hline
Interaction & Concerns compatibility with another drug/substance. & Can I drink cataflam when I drink medrol? \\
\hline
Action & Mechanism of action or physiological effect. & How does xarelto affect homeostasis? \\
\hline
Appearance & Asks about physical look (colour, shape, imprint). & What color is 30 mg prednisone? \\
\hline
Usage/Time & Best time of day to take the medicine. & When is the best time to take lotensin? \\
\hline
Stopping/Tapering & How to discontinue or taper. & How to come off citalopram? \\
\hline
Ingredient & Active ingredient(s) contained in a product. & What opioid is in the bupropion patch? \\
\hline
Action/Time & Onset/duration of drug effect. & How soon does losartan affect blood pressure? \\
\hline
Storage and Disposal & Proper storage temperature or disposal method. & In how much temp should BCG vaccine be stored? \\
\hline
Comparison & Compares two therapies/drugs. & Why is losartan prescribed rather than a calcium channel blocker? \\
\hline
Contraindication & Whether the drug is safe given allergy/condition. & If I am allergic to sulfa can I take glipizide? \\
\hline
Overdose & Consequences of taking too much. & What happens if your child ate a Tylenol tablet? \\
\hline
Alternatives & Asks for substitute medications. & What medicine besides statins lowers cholesterol? \\
\hline
Usage/Duration & How long treatment should continue. & How long should I take dutasteride? \\
\hline
Time (Other) & Other time-related effectiveness/protection questions. & How long are you protected after the Hep B vaccine? \\
\hline
Brand Names & Asks for commercial brand names. & What is brand name of acetaminophen? \\
\hline
Combination & How to combine two treatments in one regimen. & How to combine dapagliflozin with metformin? \\
\hline
Pronunciation & How to pronounce a drug name. & How do you pronounce Humira? \\
\hline
Manufacturer & Asks who makes or markets the drug. & Who makes nitrofurantoin? \\
\hline
Availability & Whether the drug is still on the market/shortages. & Has lisinopril been taken off the market? \\
\hline
Long-term-Consequences & Long-term effects of prolonged use. & What are the long-term consequences of using nicotine? \\
\hline
\end{tabular}
\label{fig:Question Type Categories}
\end{table*}
\begin{figure*}[htbp]
  \centering
  \caption{Distribution of ICD-11 Categories in PerMedCQA}
  \includegraphics[width=\linewidth]{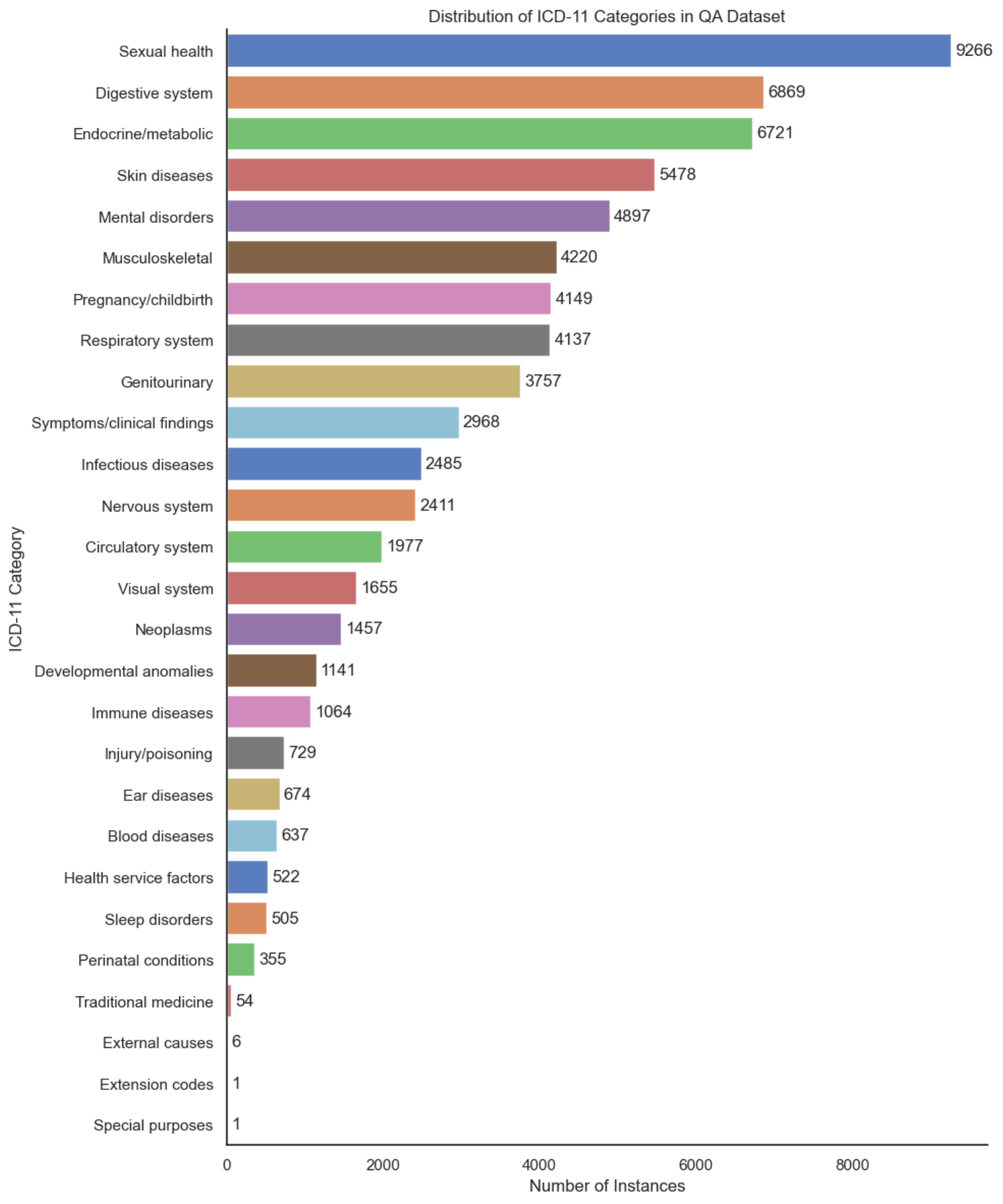}
  \label{fig:Distribution-ICD-11}
\end{figure*}
\begin{figure*}[htbp]
  \centering
  \caption{Distribution of Question Type in PerMedCQA}
  \includegraphics[width=\linewidth]{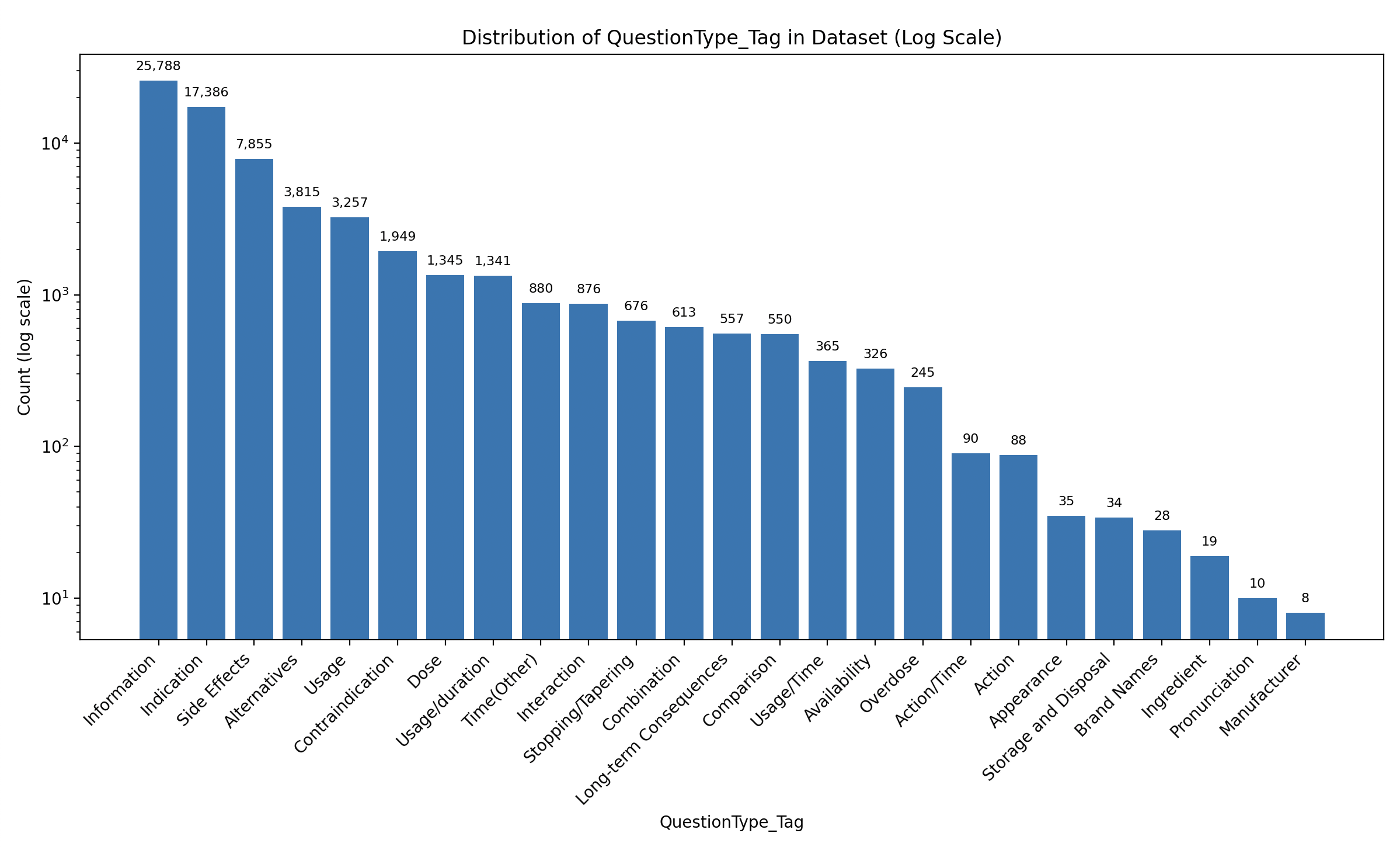}
  \label{fig:QuestionType Distribution}
\end{figure*}
\begin{figure*}[t]
\caption{Prompt description for the Question Type Tagging Task.}
\begin{tcolorbox}[
    enhanced,
    width=\textwidth,
    colback=gray!5,
    colframe=blue!75!black,
    title=\textbf{Question Type Tagging System Prompt},
    fonttitle=\bfseries,
    sharp corners=south,
    boxrule=0.6pt
  ]
\small

\textbf{Task Description}\\
You are a medical NLP annotator. Label every consumer medical question with:

\vspace{0.5em}
\begin{itemize}
  \item \texttt{question\_type\_explanation} – a concise and brief (1 sentence) justification of \textbf{why} the chosen \texttt{question\_type} is appropriate, citing question wording and any domain knowledge you considered.
  \item \texttt{question\_type} – \textbf{exactly one} of the 25 categories listed below in the table (use the label verbatim).
\end{itemize}

\end{tcolorbox}
\label{fig:Question Type Tagging Task-25}
\end{figure*}
\begin{figure*}[t]
\caption{Baseline system prompt used for all models in the default setting.}
\begin{tcolorbox}[
    enhanced,
    width=\textwidth,
    colback=gray!5,
    colframe=blue!75!black,
    title=\textbf{Baseline System Prompt (Main Prompt)},
    fonttitle=\bfseries,
    sharp corners=south,
    boxrule=0.6pt
  ]
\small

You are an experienced professional medical doctor providing reliable medical advice in fluent Persian.  
Users are reaching out through an online medical forum seeking clear, concise, and accurate answers to their health-related questions.  
Respond directly, briefly, and precisely to their inquiries.  
Always provide practical and actionable medical guidance.  
Refer users to consult a doctor only when it is strictly necessary and impossible to safely or accurately address their concerns online.  
Prioritize addressing their concerns clearly and thoroughly without unnecessary elaboration or hesitation.

\end{tcolorbox}
\label{fig:Baseline system prompt}
\end{figure*}
\begin{figure*}[t]
\caption{Role-based prompting strategy using ICD-11-conditioned specialist titles (e.g., \texttt{Psychiatrist}, \texttt{Gynecologist}).}
\begin{tcolorbox}[
    enhanced,
    width=\textwidth,
    colback=gray!5,
    colframe=blue!75!black,
    title=\textbf{Role-Based Prompt (ICD-11 Conditioned)},
    fonttitle=\bfseries,
    sharp corners=south,
    boxrule=0.6pt
  ]
\small

You are an experienced \texttt{\{role\}} providing reliable medical advice in fluent Persian.  
Users are reaching out through an online medical forum seeking clear, concise, and accurate answers to their health-related questions.  
Respond directly, briefly, and precisely to their inquiries.  
Always provide practical and actionable medical guidance.  
Refer users to consult a doctor only when it is strictly necessary and impossible to safely or accurately address their concerns online.  
Prioritize addressing their concerns clearly and thoroughly without unnecessary elaboration or hesitation.

\end{tcolorbox}
\label{fig:Role-based prompting strategy}
\end{figure*}

\begin{figure*}[t]
\caption{Translation pipeline (separate LLM inference) for leveraging English LLMs with Persian input and output.}
\begin{tcolorbox}[
    enhanced,
    width=\textwidth,
    colback=gray!5,
    colframe=blue!75!black,
    title=\textbf{Translation-Based Prompting (Three-Step Pipeline)},
    fonttitle=\bfseries,
    sharp corners=south,
    boxrule=0.6pt
  ]
\small

\textbf{Step 1: Translate Question (Persian → English)}\\
You are a professional medical translator.  
Translate the following Persian medical question into clear, neutral English.  
Only output the translation.

\bigskip
\textbf{Step 2: Answer in English}\\
You are an experienced professional medical doctor providing reliable advice in English.  
Answer directly, briefly and precisely.  
Provide practical guidance and advise an in-person visit only when absolutely necessary.

\bigskip
\textbf{Step 3: Translate Answer (English → Persian)}\\
You are a professional medical translator.  
Translate the following English answer into fluent Persian suitable for the patient.  
Only output the translation.

\end{tcolorbox}
\label{fig:Translation pipeline}
\end{figure*}

\begin{figure*}[t]
\begin{tcolorbox}[
    enhanced,
    width=\textwidth,
    colback=gray!5,
    colframe=blue!75!black,
    title=\textbf{MedJudge System Prompt},
    fonttitle=\bfseries,
    sharp corners=south,
    boxrule=0.6pt
  ]
\small

You are \texttt{"Med-Judge"}, an objective medical QA grader responsible for evaluating how closely a Model-generated answer aligns with an Expert-provided answer (the gold-standard).

\bigskip
\textbf{Strict Guidelines for Evaluation:}

\begin{enumerate}
    \item \textbf{Exclusivity:}  
    Base your judgment \textbf{only on the Expert's provided answer}, and never rely on your own medical knowledge or external resources.

    \item \textbf{Step-by-step and Explainable Evaluation:}
    \begin{itemize}
        \item Provide a brief and concise comparison analysis (\texttt{brief\_analysis}), clearly explaining your evaluation.
        \item Clearly identify the key medical facts, statements, or recommendations in both Expert and Model answers.
        \item List explicitly any critical facts appearing \textbf{only} in the Expert’s answer (\texttt{key\_missing\_facts}) or \textbf{only} in the Model’s answer (\texttt{key\_extra\_facts}).  
        These should be keywords in Persian (Farsi).
    \end{itemize}

    \item \textbf{Labeling Correctness:}  
    Categorize the answer strictly according to these definitions:
    \begin{itemize}
        \item \texttt{correct} – Answers have essentially identical meaning with no clinically meaningful differences.
        \item \texttt{partially\_correct} – Answers differ slightly without critical differences affecting clinical understanding or advice.
        \item \texttt{incorrect} – Answers substantially differ, significantly impacting clinical meaning or completeness.
        \item \texttt{contradictory} – Model explicitly contradicts Expert’s answer or provides opposite medical advice.
    \end{itemize}

    \item \textbf{Coverage Analysis:}  
    Evaluate factual coverage using:
    \begin{itemize}
        \item \texttt{equal} – Both answers contain exactly the same key facts.
        \item \texttt{model\_subset} – Model answer misses one or more critical facts present in the Expert answer.
        \item \texttt{expert\_subset} – Model answer includes additional key facts not found in the Expert’s answer.
        \item \texttt{overlap\_none} – No meaningful shared key facts between answers.
    \end{itemize}

    \item \textbf{Clinical Impact Estimation:}  
    Judge how the differences could clinically affect a patient’s safety or care using:
    \begin{itemize}
        \item \texttt{negligible}, \texttt{moderate}, \texttt{significant}, \texttt{critical}
    \end{itemize}

    \item \textbf{Confidence Rating:}  
    Express your certainty level:
    \begin{itemize}
        \item \texttt{high}, \texttt{medium}, \texttt{low}
    \end{itemize}
\end{enumerate}

\bigskip
\textbf{Attention:}
\begin{itemize}
    \item Be strict and literal in comparing the two answers.
    \item Do not assume or infer correctness beyond the Expert answer.
    \item Be concise and structured. Avoid vague commentary.
\end{itemize}

\bigskip
\textbf{Output Format:}  
Return \textbf{only} a valid JSON object matching the schema.
\end{tcolorbox}
\caption{MedJudge system prompt with structured grading rubric for model evaluation.}
\label{fig:MedJudge system prompt}
\end{figure*}

\begin{figure*}[t]
\begin{tcolorbox}[
    enhanced,
    width=\textwidth,
    colback=gray!5,
    colframe=blue!75!black,
    title=\textbf{MedJudge User Prompt Template},
    fonttitle=\bfseries,
    sharp corners=south,
    boxrule=0.6pt
  ]
\small

\textbf{Question:}  
\{\texttt{question}\}

\bigskip
\textbf{Expert Answer:}  
\{\texttt{expert\_answer}\}

\bigskip
\textbf{Model Answer:}  
\{\texttt{model\_answer}\}

\end{tcolorbox}
\caption{Template used for submitting evaluation tasks to MedJudge. Placeholders are replaced at runtime.}
\label{fig:User prompt}
\end{figure*}

\begin{figure*}[t]
    \centering
    \caption{Human Evaluation}
    \includegraphics[width=\linewidth]{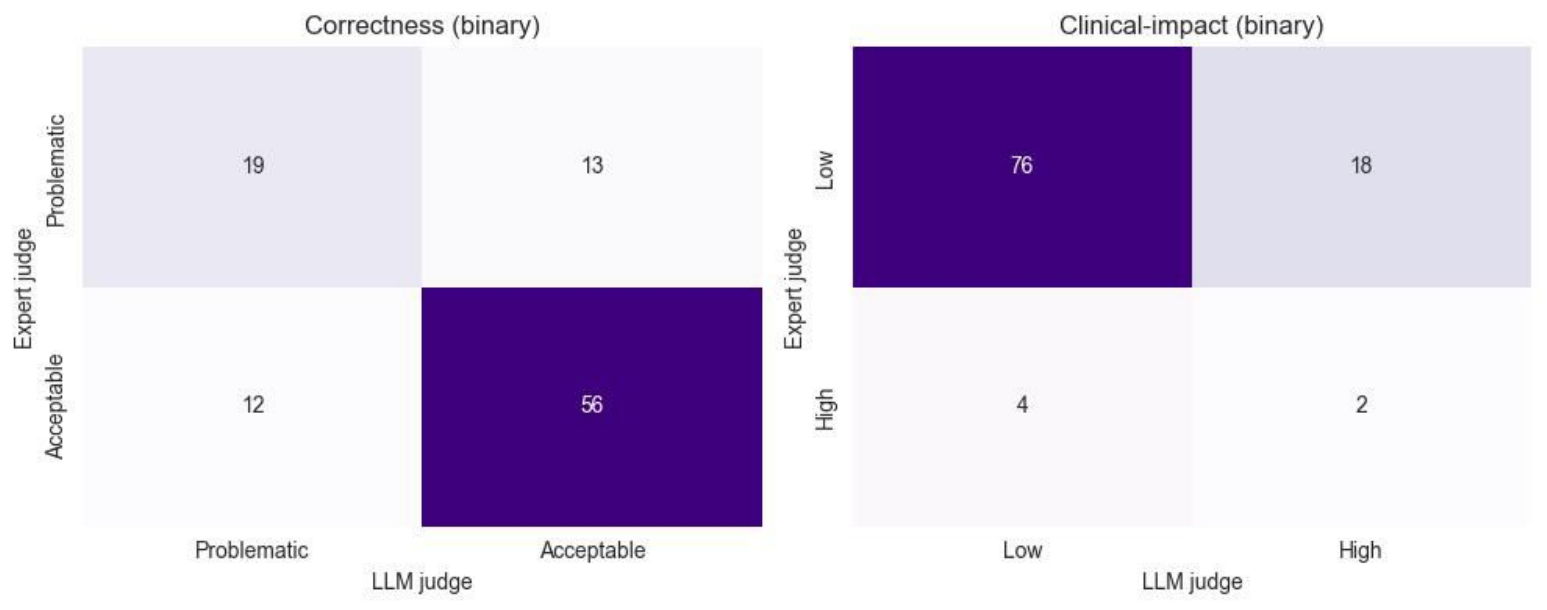}
    \label{fig:Human Evaluation}
\end{figure*}
\begin{figure*}[t]
    \centering
    \caption{Correctness for baselines and advanced Prompting}
    \includegraphics[width=\linewidth]{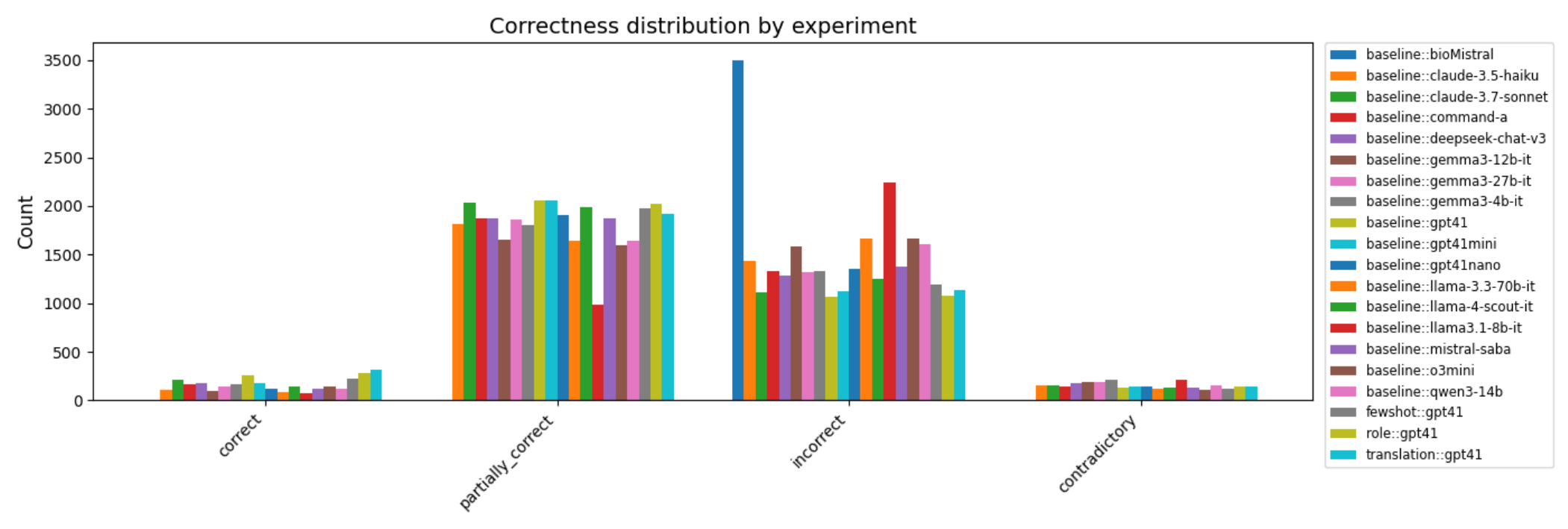}
    \label{fig:baselines_correctness}
\end{figure*}
\begin{figure*}[t]
    \centering
    \caption{Coverage for baselines and advanced Prompting}
    \includegraphics[width=\linewidth]{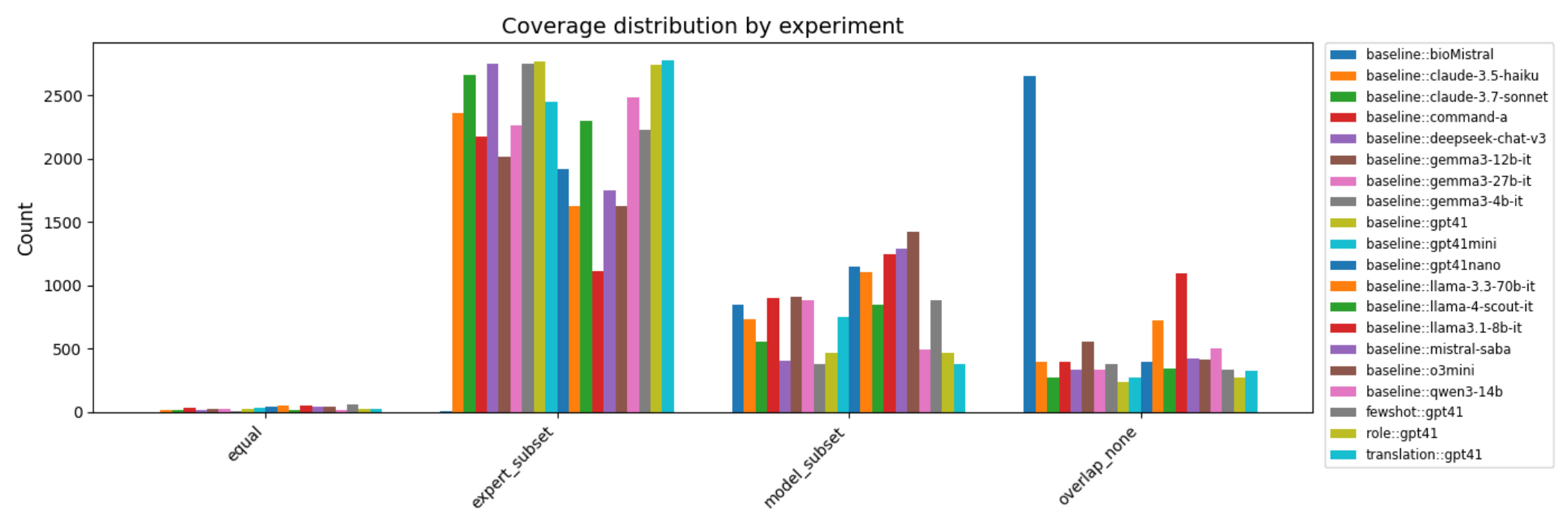}
    \label{fig:baselines_coverage}
\end{figure*}
\begin{figure*}[t]
    \centering
    \caption{Critical impact for baselines and advanced Prompting}
    \includegraphics[width=\linewidth]{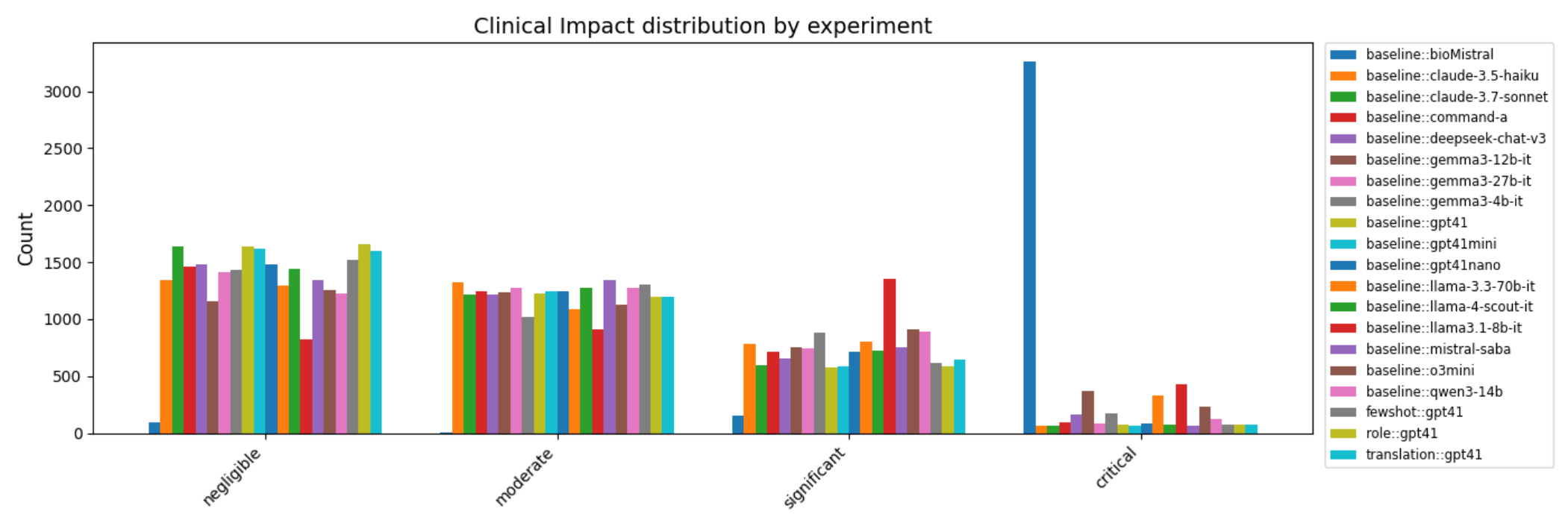}
    \label{fig:baselines_clinical_imapcat}
\end{figure*}

\end{document}